%% file: main.tex
\pdfoutput=1

\documentclass[11pt]{article}

\usepackage[final]{acl}

\usepackage{times}
\usepackage{latexsym}

\usepackage[T1]{fontenc}
\usepackage[utf8]{inputenc}

\usepackage{microtype}
\usepackage{inconsolata}

\usepackage{amssymb}
\usepackage{amsmath} 

\usepackage{graphicx}
\usepackage{tabularx}

\usepackage{booktabs}
\usepackage{multirow}

\usepackage{tikz}
\usetikzlibrary{arrows, calc, positioning, shapes}

\input{sections/001-Meta}

\begin{document}
\maketitle

\input{sections/002-Abstract}

\input{sections/101--Instruction}

\input{sections/102--Related-Work}
\input{sections/103-Formalization}

\input{sections/104--TWON-Case-Study}
\input{sections/105--Experiments}
\input{sections/106--Results}
\input{sections/107--Conclusion}
\input{sections/201--Postscript}


\bibliography{references}


\end{document}

%% file: sections/001-Meta.tex
\title{
    Don't Trust Generative Agents to Mimic Communication on \\
    Social Networks Unless You Benchmarked their Empirical Realism
}

\author{
    Simon Münker \\
    Trier University \\
    \texttt{muenker@uni-trier.de} 
\And
  Nils Schwager \\
  Trier University \\
  \texttt{schwager@uni-trier.de}
\And
  Achim Rettinger \\
  Trier University \\
  \quad\texttt{rettinger@uni-trier.de}
}

%% file: sections/002-Abstract.tex
\begin{abstract}
The ability of Large Language Models (LLMs) to mimic human behavior triggered a plethora of computational social science research, assuming that empirical studies of humans can be conducted with AI agents instead. Since there have been conflicting research findings on whether and when this hypothesis holds, there is a need to better understand the differences in their experimental designs. We focus on replicating the behavior of social network users with the use of LLMs for the analysis of communication on social networks. First, we provide a formal framework for the simulation of social networks, before focusing on the sub-task of imitating user communication. We empirically test different approaches to imitate user behavior on $\mathbb{X}$ in English and German. Our findings suggest that social simulations should be validated by their empirical realism measured in the setting in which the simulation components were fitted. With this paper, we argue for more rigor when applying generative-agent-based modeling for social simulation.
\end{abstract}

%% file: sections/101--Instruction.tex
\section{Introduction}
Social media platforms face mounting regulatory pressure worldwide. Australia and Spain are implementing age restrictions for users under 16\footnote{Australia approves social media ban on under-16 \url{https://www.bbc.com/news/articles/c89vjj0lxx9o}}, while several countries have banned TikTok due to concerns about misinformation \cite{basch2021global} and harmful content \cite{weimann2023research}. The EU's Digital Service Act mandates that large platforms "identify, analyze and assess any systemic risks [...] from the use made of their services," including "actual or foreseeable negative effects on civic discourse."

Meeting these regulatory requirements demands robust quantitative evidence about platform risks—particularly the role of algorithmic mechanisms like ranking and recommendation systems. Yet obtaining such evidence remains challenging in computational social and communication science \cite{wang2016understanding}. The fundamental problem is causal inference: determining whether alternative platform designs would produce different, potentially more favorable outcomes. This question is best addressed through simulation, which enables testing counterfactual scenarios without exposing real users to potentially harmful conditions.

The validity of simulation-based insights depends critically on how faithfully simulations represent reality. Recent advances in Large Language Models (LLMs) have enabled more sophisticated agent communication in social simulations \cite{chuang2024simulating}, surpassing the realism of traditional symbolic agent languages. However, this increased sophistication raises a crucial question: Can LLM-based agents replicate human social behavior with sufficient fidelity to produce scientifically robust insights?

Answering this question requires systematic experimental designs that improve reproducibility and provide quantitative measures of \textit{empirical realism}, the degree to which simulated behaviors match observed human behaviors. Without such validation, simulation findings lack the credibility needed to inform policy decisions or scientific understanding.

\subsection*{Contributions}
Our work advances the rigor of LLM-based social network simulation through three key contributions:

\paragraph{Framework}
We formalize social networks to (i) standardize their simulation and (ii) enable systematic benchmarking of empirical realism through quantifiable metrics.

\paragraph{Case Study}
We instantiate this framework with multiple approaches for imitating user communication on $\mathbb{X}$, including posting, replying, and reply-likelihood prediction tasks across English and German datasets.

\paragraph{Validation}
We benchmark empirical realism across tasks, languages, and modeling approaches, revealing when and how LLMs can faithfully replicate user behavior, and crucially, when they cannot.

%% file: sections/102--Related-Work.tex
\section{Related Work}

\subsection{From Symbolic Agents to Neural Language Models}
Social simulation has evolved from symbolic \cite{o1998fipa}, rule-based agents to data-driven neural approaches. Traditional multi-agent systems \cite{weiss1999multiagent} employed explicit cognitive models with deductive reasoning, while game-theoretic frameworks relied on rational choice assumptions. These approaches prioritized theoretical interpretability and control over predictive accuracy, a reasonable trade-off when modeling capabilities were limited.

Our work diverges fundamentally from this tradition. Rather than constructing agents from theoretical principles, we aim to replicate observed communication patterns as precisely as possible. This shifts the goal from explanation to imitation, and from symbolic representations to natural language generation. The question is not whether agents follow theoretically motivated rules, but whether they produce outputs statistically indistinguishable from real users.

\subsection{LLMs as Human Simulacra: Promise and Limitations}
The use of LLMs as human surrogates began with believable non-player characters in game environments \cite{park2023generative}, demonstrating superficially authentic behavior. This success sparked broader interest in using LLMs as synthetic participants across social sciences \cite{argyle2023out, chuang2024simulating, papachristou2024network}. Proponents argue that appropriate prompting enables models to accurately emulate response distributions from diverse demographic subgroups \cite{argyle2023out, chang2025llms}.

However, growing evidence challenges these optimistic claims. LLM-generated synthetic opinions reveal fundamental limitations of prompting-based alignment. While persona prompting approximates US opinion distributions, performance degrades substantially for non-English populations \cite{argyle2023out, ma2024algorithmic, munker2025political, munker2025cultural}. This suggests prompting privileges cultural contexts embedded in training data rather than reliably grounding outputs in external reality.

Moreover, prompting produces unstable representations. Minor prompt variations dramatically affect behavioral outputs \cite{ma2024algorithmic}, and RLHF training compounds this instability \cite{tjuatja2024llms}. These findings indicate that prompting inadequately controls model behavior, failing to consistently activate intended knowledge representations.

\subsection{Methodological and Ethical Concerns}

Beyond technical limitations, using LLMs as human proxies raises serious methodological questions. Training data biases can systematically misrepresent marginalized groups \cite{abid2021persistent, hutchinson2020social}. The disembodied nature of LLMs means they lack grounding in physical reality, cultural contexts, physical environments, and interpersonal relationships—that fundamentally shapes human cognition and decision-making \cite{hussein2012sapir, argyle2023out}.

Ethically, deploying LLMs as participant replacements raises concerns about informed consent, potential misuse, and oversimplification of complex human behaviors. These concerns intensify when researchers make causal claims based on simulated data without validating empirical realism.

\paragraph{Our Position} We argue that LLM-based simulation can contribute meaningfully to social science, but only when coupled with rigorous validation against real-world data. Claims about simulation validity must be empirically demonstrated, not assumed. This paper provides a framework for such validation.

%% file: sections/103-Formalization.tex
\section{Modeling Online Social Networks}
\label{sec:OSN-Model}
\begin{figure*}
    \centering
    \includegraphics[width=1\linewidth]{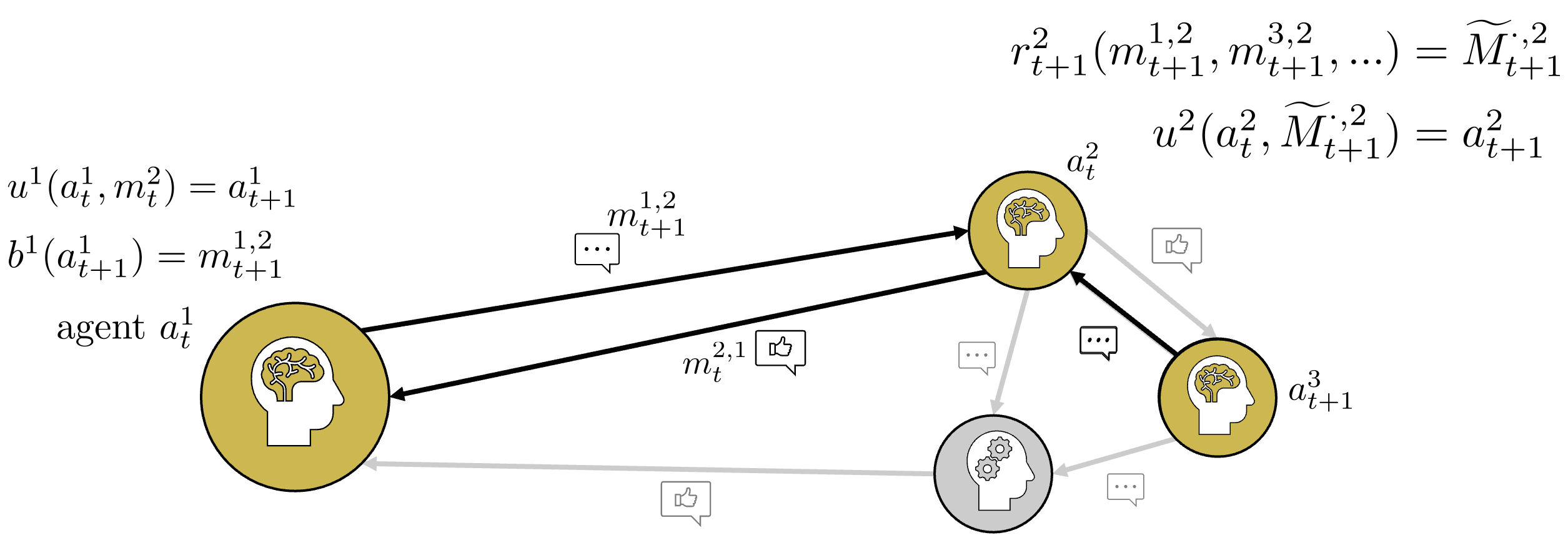}
    \caption{
        Illustrating example:
        Agent ($a^2$) sends a message to agent ($a^1$), triggering a model update and reply generation. Agent ($a^2$) receives a curated list of incoming messages at $t+1$ and updates its model accordingly. This illustrates the interplay between agent behavior ($b$) and network mechanics ($r$).
    }
    \label{fig:agent-formalization}
\end{figure*}

We present a formal framework for Twins of Online Social Networks (TWONs) that captures both user behavior (Sect.~\ref{subsec:ModelingAgents}) and platform mechanics (Sect.~\ref{subsec:ModelingNetworks}). This framework enables systematic comparison between simulated and observed behaviors. Figure~\ref{fig:agent-formalization} illustrates key concepts.

\subsection{Modeling Agents}
\label{subsec:ModelingAgents}
Let $A$ be a set of agents, i.e., social media users, where $a_t^i$ is the state of the $i$-th agent at time $t$.\footnote{The internal state of an agent can be represented by symbolic logic, machine learned functions, cognitive architecture or any other agent formalism that is capable of perceiving its environment and acting upon that environment.}

$M$ is a set of messages, where $m_t^{(i,j)}$ is a message from agent $i$ to agent $j$ at time $t$. Consequently, $m_t^{(i,\cdot)}$ indicates a message that agent $i$ transmits to all other agents and $M_t^{(\cdot, i)}$ indicates all messages sent to agent $i$. Accordingly, $M_t^i$ indicates all messages that agent $i$ sends and receives at time $t$.

The state $a_t^i$ of the $i$-th agent at time $t$ is defined by the discourse it has been involved in up to this point: $(M_0^i, M_1^i,... , M_t^i)$. The agent's state is updated according to the function $u^i(a_{t}^i, M_t^i) = a_{t+1}^i$ based on its previous state and the messages sent and received since the last point in time.

The communicative behavior $b^i(\cdot)$ of the agent $i$ is a function that maps its current state $a_t^i$ to the messages $m_{t+1}^{(i,j)}$ he is sending next to agents $j$: $b^i_t(a_t^i) = M_{t}^{(i,j)}$.

\subsection{Modeling Network Mechanics}
\label{subsec:ModelingNetworks}
The mechanics of the communication channel (here, an online social network) is defined by a function $r^i_t(\cdot)$ that adapts the agent $i$' to the incoming messages $M_t^{(\cdot, i)}$ to $\widetilde{M}_t^{(\cdot, i)}$, returning a manipulated list of messages. Standard manipulations of the list are filtering or ranking, but could also include the manipulation of content or the addition of messages to the list that have not been directly sent between agents. Note that this function can be personalized and be adapted over time.

In general, the messages agent $i$ receives at time $t$ are given by:



$$ r_t^i( b^{j}_t(a_t^j)) = \widetilde{M}_{t}^{(\cdot, i)} \quad \forall j \in A_t^j  $$


which in turn triggers agent $i$s response 

$$ b^i_{t+1}( u^i(a^i_t, \widetilde{M}_{t}^{(\cdot, i)})) = M_{t+1}^{(i,\cdot)}$$

\section{Simulating TWONs}
\label{sec:replication}
Based on the formal framework introduced in Sect.~\ref{sec:OSN-Model} we define the required tasks to construct a TWON by replicating its behavior at the user (Sect.~\ref{subsec:replicatingUsers}) and system level (Sect.~\ref{subsec:replicatingUsers}). Finally, this allows us to check the empirical realism of such TWONs (Sect.~\ref{subsec:simulation}).

\input{figures/methods.tasks}

\paragraph{Imitating User Behavior}
\label{subsec:replicatingUsers}
The task of machine learning an agent-based simulation of human social media communication behavior at user level is to estimate the function $b^i(\cdot)$ of each agent $i$, which given real-world observations of discourses $(M_0^i, M_1^i,... , M_t^i)$ predicts its next message $m_{t+1}^i$. Thus, the objective can be formulated as minimizing a loss function 

$$\textrm{min\ } L_{b}(\widehat{m}_{t+1}^i, m_{t+1}^i) \mapsto \mathbb{R}$$

that compares the predicted $\widehat{m}_{t+1}^i$ to its observed value.\footnote{Any text comparison metric, like BLEU, cross-entropy, but also higher level metrics could be applied here. In addition, other metrics can be used to measure the predictive accuracy of discrete actions such as "liking".}

\paragraph{Replicating Network Mechanics}
\label{subsec:replicatingMechanics}
The task of replicating the system-level mechanics of a social network from observations is to estimate the function $r^i$ of each agent $i$, given its observed messages $(\widetilde{M}_1^i, \widetilde{M}_1^i,... , \widetilde{M}_t^i)$.

This objective can be formulated as minimizing the loss 

$$\textrm{min\ } L_{r}(\widehat{M}_t^{(\cdot, i)}, \widetilde{M}_t^{(\cdot, i)}) \mapsto \mathbb{R}$$

that compares the predicted $\widehat{M}_t^{(\cdot, i)}$ with its observed value.

\paragraph{Simulating Social Networks}
\label{subsec:simulation}
Finally, a simulation task for investigating user communication on social networks can be stated as follows: Given (an estimation of) $b$ and $r$ generate a discourse $M_{t+1}, ..., M_{t+n}$ and evaluate it according to a discourse metric $q$, for example, the degree of outrage or hate speech, mapping 
$$q(M_{t+1}, ..., M_{t+n}) \mapsto \mathbb{R}$$

Any findings obtained by $q$ can be put into perspective by $L_{b}$ and $L_{r}$ as they quantify the empirical realism of the simulation of a social network and consequently qualify the validity of $q$. $L_{b}$ and $L_{r}$ can be interpreted as a confidence score for any empirical findings obtained by simulating social networks and should be reported together with empirical findings obtained by simulation.

%% file: figures/methods.tasks.tex
\begin{table*}[ht]
	\centering
    \small
	\begin{tabular}{l | llllll}
		\toprule
		Behavior $b$ & 
        Agent $a^i$  & 
        Task Type & 
        Input $\widetilde{M}_{t}^{(\cdot, i)}$ & 
        Output $m_{t+1}^{(i,\cdot)}$ & 
        Loss $L_b$ &
        Metric(s) $q$ \\
		\midrule
		Posting & Creator & Generative  & Topic  & Free Text & Cross Entropy & BLEU, ngram,... \\
		Replying & Reactor & Generative  & Post   & Free Text & Cross Entropy & BLEU, ngram,... \\
		Replying Likelihood & Reactor & Probability & Post & Interval [0,1] & Binary Cross Entropy & F1, Acc. \\
		\bottomrule
	\end{tabular}
    \vspace{0.1cm}
	\\
	\caption{Communicative behaviors modeled, imitated and benchmarked in our empirical study with respect to their empirical realism, as definied in our formal framework.}
	\label{tab:methods:task}
\end{table*}

%% file: sections/104--TWON-Case-Study.tex
\section{Case Study Design: Imitating $\mathbb{X}$ Users}
\label{sec:twon}

We now instantiate our framework using data from $\mathbb{X}$. Our implementation focuses on user behavior function $b^i$ while acknowledging simplifications in other framework components. We estimate communicative behavior by minimizing $L_b(\widehat{m}_{t+1}, m_{t+1})$ across three action types: posts, replies, and reply-likelihood.

The imitation of agent behavior $b$ is further separated into distinct subtasks: \emph{Posting} behavior is characteristic for the content creator user group and the \emph{Replying} behavior for reactive users. In our data, the former category includes politicians, policy makers, and institutional news outlets, characterized by their role in initiating discussions and setting conversational agendas. The latter comprises individual users whose primary mode of engagement involves responding to and amplifying existing content through platform-specific mechanics such as replies, retweets, and reactions. This asymmetric structure reflects the empirically observed power-law distribution of participation in social networks, where a small percentage of users generate most of the original content \citep{carron2014describing, sun2014understanding}.

An overview of the different agent behaviors that we empirically evaluate in the next sections is provided in Table~\ref{tab:methods:task} and is described below.

\paragraph{Imitating Posting Behavior}
We model party-specific content generation patterns through the extracted topics and the public information (name and party affiliation) of the individuals - in our case politicians. Given a specified topic, the model learns to generate content that reflects both the broad party stance and individual variations in expression style.

\paragraph{Imitating Replying Behavior}
The reply generation system operates on a context-aware framework that processes historical interaction patterns. The historical interaction context is represented as a sequence of <Post, Reply> pairs, encoded as few shot examples during the alignment process. The model maintains consistent user behavior by implementing a constraint mechanism that prevents top-level post-generation, restricting outputs to reply-only contexts. 

\paragraph{Estimating Replying Likelihood}
The goal of this task is to predict whether a synthetic social network user would respond to a post based on their history of interaction. The system processes historical interactions and a current post using a fine-tuned BERT model \cite{zhang2023twhin} to generate embeddings by calculating the pooled mean of its input representations. Historical interactions and the post are separately processed through quadratic linear layers with ReLU activation, maintaining dimensions related to historical data and encoder size. The post embedding is repeated to match each historical entry and combined using element-wise multiplication with the history representation, establishing an interaction between past patterns and current post features. These combined representations are fed into a classification layer, producing a prediction score between 0 and 1 to indicate the likelihood of a synthetic user replying to the post.

%% file: sections/105--Experiments.tex
\section{Fitting Agents to User Data from $\mathbb{X}$}
\label{sec:fitting}
In this Section, we describe the data set used to train and evaluate our approaches for imitating the types of user behavior described in the previous Section. Furthermore, the machine learning processes for fitting the models to the data is described.

\paragraph{Dataset}
\label{subsec:data}
Our experiments are based on two datasets, English and German, collected from $\mathbb{X}$. The datasets are collected around keywords concerning the political discourses in the US and Germany during the first half of 2023. The samples contain two types of content: a) Tweets (posts) from delegates of the national parliament concerning political decisions – DE: $155.000$, EN: $930.000$ and b) replies from regular $\mathbb{X}$ users towards these decisions or opinions – DE: $185.000$, EN: $17.800.000$. In addition, we labeled the topics of posts and replies using zero-shot prompt classification with Llama 3.1 70B \cite{zheng2023judging}. This yields two distinct agent types, the a) political actor for posts and the b) unspecified $\mathbb{X}$ user for replies, both across two languages. 

\paragraph{Preprocessing}
\label{subsec:preprocessing}
We apply moderate content filtering to enhance the quality of the selected data:

\emph{External Source:} We remove samples that include URLs to external sources. Although this significantly reduces the number of samples, our models cannot access this information during the training and inference process.

\emph{Retweets:} Tweets starting with \textit{RT} mark a shared post. The author does not add new content, but replicates the original text and shares it with their followers. We exclude these samples because we cannot verify if the shared content is in line with the retweeting author's opinion. 

\emph{Content Length:} We aim to focus on content that contains arguments or opinions, particularly given the later applied discourse metrics. Thus, we remove samples with less than $32$ characters based on our heuristic assumption that a specified length is necessary to convey an opinion.

After the initial pre-processing, we further reduce both datasets to retrieve only the most active users measured by the number of posts and replies. This is required for modeling both tasks on a user-based level by providing user examples (few-shot) during prompting.

\paragraph{Content Generation Alignment Pipeline}
\label{subsec:training}
We train two model adapters on top of Llama-3.2-3B-Instruct \citep{dubey2024llama} using the supervised fine-tuning (SFT) paradigm through the transformer reinforcement learning (TRL) library using the default pipeline. The trained adapters are unmodified LoRA \citep{hu2021lora} matrices provided by the parameter-efficient fine-tuning (PEFT) package. We optimize for two objectives (posting and replying) with contents from the original users as the labeled data.

\paragraph{Behavior Likelihood Alignment Pipeline}
For the decision task, we train our dual input model using embeddings generated by a frozen BERT-based encoder that was pretrained to capture semantic representations of tweets \citep{zhang2023twhin}. We optimize with AdamW \citep{loshchilov2017fixing} using the same data set as for the generative tasks, making the assumption that when a user has not commented on tweets in our collection, this represents an active decision not to reply. To maintain balance, we sample an equal number of negative and positive instances for both languages.

\paragraph{Data and Code Availability}
The complete technical pipeline for this study is publicly available on GitHub at \url{https://anonymous.4open.science/r/TWON-Agents-D0E2}. This repository includes the source code for all the data processing, training, and evaluation steps described in this paper. The raw and processed datasets used in this study will be made available after acceptance.


%% file: sections/106--Results.tex
\section{Evaluating the Empirical Realism of Imitated $\mathbb{X}$ Users}
We evaluated the approaches for the different imitation tasks as described in Sect.~\ref{sec:twon} and Sect.~\ref{sec:fitting} to assess their empirical realism.

\include{figures/results.generative.tex}

\subsection{Evaluation of Imitated Posting and Replying Behavior}
The text-generation-based tasks are evaluated with two different approaches, in two different languages, using three text comparison metrics (see.~Table~\ref{tab:results:generative}).

\paragraph{In-context vs Fine-tuning}
An initial observation is that BLEU scores \cite{papineni2002bleu} in all in-context settings are low. This is also true for the n-gram metrics. However, the fine-tuning-based approach shows significant improvements specifically in the English reply task, but also for uni-gram matches and the overall length ratio. Beyond such token-overlap-metrics, we tested semantic similarity metrics: we embedded all samples using $\mathbb{X}$-fine-tuned BERT \citep{zhang2023twhin} and calculated the distance between generations and original samples. Again, the results generated from the fine-tuned model show a significantly shorter distance to its original counterpart compared to the in-context model. The correlation between the predicted emotions in the original and generated samples using a BERT-based multi-label emotion classifier \citep{barbieri2020tweeteval} is less clear. The fine-tuned approach only marginally outperforms the prompt-only approach on the posting task. 

However, the English fine-tuned reply generation task showed by far the best performance with significant improvements in all metrics. The BLEU score increased substantially from 0.019 to 0.239, uni-gram precision improved from 0.167 to 0.387, and the embedding distance was reduced significantly from 2.006 to 1.427. The model achieved a near-perfect length ratio of 1.003 and demonstrated strong TweetEval correlations across all categories.

\paragraph{English vs German}
In general, we see the same improvements through fine-tuning compared to in-context with respect to the German data in the post-generation task: The BLEU score increases and uni-gram precision improves. Also, the embedding distance was reduced from 2.530 to 1.499, and the length ratio moved closer to 1.0 after fine-tuning (1.023 vs 1.142). However, compared to English, the TweetEval metrics proved generally unreliable for German content, showing inconsistent patterns with high standard deviations. This suggests fundamental limitations in applying English-trained evaluation metrics to German content. Similarly, the German reply generation (history-based) task shows low empirical realism on all metrics compared to English and regardless of fine-tuning. There were minimal improvements in BLEU scores from 0.005 to 0.021, with high variance in metrics, exemplified by the bi-gram standard deviation of 0.044. The embedding distance remains high even after fine-tuning at 2.891.

\subsection{Evaluation of Replying Likelihood Behavior}
Similar to our evaluation of the generative tasks, we observe substantial performance disparities between English and German models (Table \ref{tab:results:probability:test}). Even on the German training set, our model achieves only moderate alignment with an F1-Score of $0.666$, in stark contrast to the English experiment, where the prediction scores are almost perfect. This performance gap suggests that the BERT encoder employed in our study represents English samples with semantically richer embeddings than their German counterparts. 

\input{figures/results.probability.test}

\subsection{Key Findings}
Several key findings emerge from this analysis. First, language-specific performance shows that English models significantly outperform German models across all metrics, with TweetEval metrics proving more reliable for English content while German models show limited success, particularly in reply generation. Second, task-specific patterns reveal that fine-tuning shows consistent improvements for post-generation in both languages. However, reply generation varies dramatically between languages, with English reply generation achieving remarkably high scores. Third, the impact of data volume is evident, with better performance in English likely due to larger LLM pre-training datasets. Fourth, in terms of evaluation metrics, BLEU scores and embedding distances provide consistent signals, TweetEval metrics are only meaningful for English content, and the length ratio serves as an intuitive indicator of style authenticity.

\subsection{Imitation Limitations}
\label{subsec:ImiLimi}
The stark contrast between English and German model performance, particularly in reply generation tasks, underscores the importance of data volume and language-specific considerations when utilizing generative-agent-based modeling for imitating user behavior on social networks. The success of the English models demonstrates the potential of the approach, while the limitations faced by the German models highlight that they require additional training and optimizations to provide robust levels of empirical realism.

Our experimental setup has further limitations regarding both data selection and the system's capacity to generate discourses containing well-structured standpoints and arguments. Although our models demonstrate such capabilities under certain conditions, their performance characteristics primarily reflect the behavioral patterns of the most active users within our dataset, thus mirroring the distinctive communication dynamics observed in the selected $\mathbb{X}$ community. This sampling bias raises questions about the generalizability of our findings in different contexts of social networks and user populations.

%% file: figures/results.generative.tex
\begin{table*}[ht]
	\renewcommand{\arraystretch}{1.15}
	\centering
	\begin{tabular}{l | rr | rr}
		\toprule
		& \multicolumn{4}{c}{\textbf{English}} \\
        		\midrule
		& \multicolumn{2}{c|}{\textbf{Post (persona-based)}} & \multicolumn{2}{c}{\textbf{Reply (history-based)}} 
		\\
		& in-context & fine-tuned                   
        & in-context & fine-tuned             
		\\
		\midrule
        
		\textbf{BLEU} $^a$
		& 0.015 ($\pm$ 0.002)             & \textbf{0.085} ($\pm$ 0.017)          
        & 0.019 ($\pm$ 0.005)             & \textbf{0.239} ($\pm$ 0.042) 
		\\
		\textbf{unigram} $^b$
		& 0.157 ($\pm$ 0.005)             & \textbf{0.294} ($\pm$ 0.016)          
        & 0.167 ($\pm$ 0.008)             & \textbf{0.387} ($\pm$ 0.041) 
		\\
		\textbf{bigram} $^b$
		& 0.022 ($\pm$ 0.002)             & \textbf{0.095} ($\pm$ 0.015)          
        & 0.021 ($\pm$ 0.004)             & \textbf{0.243} ($\pm$ 0.047) 
		\\
		\textbf{length ratio} $^c$ 
		& 1.632 ($\pm$ 0.053)             & \textbf{1.001} ($\pm$ 0.054)          
        & 1.324 ($\pm$ 0.058)             & \textbf{0.916} ($\pm$ 0.040) 
		\\
		\midrule
		\textbf{TweetEval} $^d$ & & & &
		\\
		\quad topics
		& 0.587 ($\pm$ 0.189)             & 0.645 ($\pm$ 0.207)          
        & 0.337 ($\pm$ 0.216)             & 0.536 ($\pm$ 0.211)  
		\\
		\quad emotions 
		& 0.437 ($\pm$ 0.108)             & \textbf{0.562} ($\pm$ 0.105)          
        & 0.336 ($\pm$ 0.117)             & \textbf{0.479} ($\pm$ 0.107) 
		\\
		\quad sentiment 
		& 0.435 ($\pm$ 0.073)             & \textbf{0.619} ($\pm$ 0.083)          
        & 0.386 ($\pm$ 0.113)             & \textbf{0.586} ($\pm$ 0.087) 
		\\
		\quad offensive 
		& 0.491 ($\pm$ 0.061)             & \textbf{0.563} ($\pm$ 0.050)          
        & 0.379 ($\pm$ 0.098)             & \textbf{0.567} ($\pm$ 0.079) 
		\\
		\quad hate 
		& 0.327 ($\pm$ 0.178)             & 0.200 ($\pm$ 0.085)          
        & 0.231 ($\pm$ 0.191)             & \textbf{0.502} ($\pm$ 0.257) 
		\\
		\quad irony 
		& 0.190 ($\pm$ 0.085)             & \textbf{0.353} ($\pm$ 0.085)          
        & 0.169 ($\pm$ 0.120)             & \textbf{0.377} ($\pm$ 0.112) 
		\\
		\midrule
		\textbf{embed. dist.} $^e$ 
		& 2.018 ($\pm$ 0.101)             & \textbf{1.461} ($\pm$ 0.085)          
        & 2.006 ($\pm$ 0.162)             & \textbf{1.427} ($\pm$ 0.091) 
		\\
        \midrule
        		& \multicolumn{4}{c}{\textbf{German}} 
		\\

		\midrule
		\textbf{BLEU} $^a$
		& 0.009 ($\pm$ 0.002)             & \textbf{0.025} ($\pm$ 0.007) 
        & 0.005 ($\pm$ 0.001)             & 0.021 ($\pm$ 0.031) 
		\\
		\textbf{unigram} $^b$
		& 0.176 ($\pm$ 0.010)             & \textbf{0.238} ($\pm$ 0.016) 
        & 0.099 ($\pm$ 0.006)             & 0.148 ($\pm$ 0.050) 
		\\
		\textbf{bigram} $^b$
		& 0.016 ($\pm$ 0.001)             & \textbf{0.035} ($\pm$ 0.003) 
        & 0.006 ($\pm$ 0.003)             & 0.032 ($\pm$ 0.044) 
		\\
		\textbf{length ratio} $^c$ 
		& 1.142 ($\pm$ 0.044)             & \textbf{1.023} ($\pm$ 0.049) 
        & 1.961 ($\pm$ 0.234)             & \textbf{0.785} ($\pm$ 0.117) 
		\\
		\midrule
		\textbf{TweetEval} $^d$ & & & &
		\\
		\quad topics
		& 0.238 ($\pm$ 0.300)             & 0.340 ($\pm$ 0.301)          
        & 0.056 ($\pm$ 0.225)             & 0.039 ($\pm$ 0.225) 
		\\
		\quad emotions 
		& 0.123 ($\pm$ 0.278)             & 0.220 ($\pm$ 0.251)          
        & 0.060 ($\pm$ 0.215)             & 0.125 ($\pm$ 0.261) 
		\\
		\quad sentiment 
		& 0.079 ($\pm$ 0.276)             & 0.255 ($\pm$ 0.200)          
        & 0.066 ($\pm$ 0.196)             & 0.246 ($\pm$ 0.264) 
		\\
		\quad offensive 
		& 0.104 ($\pm$ 0.215)             & 0.095 ($\pm$ 0.239)          
        & 0.141 ($\pm$ 0.200)             & 0.278 ($\pm$ 0.350) 
		\\
		\quad hate 
		& 0.234 ($\pm$ 0.340)             & 0.477 ($\pm$ 0.247)          
        & 0.163 ($\pm$ 0.212)             & 0.044 ($\pm$ 0.160) 
		\\
		\quad irony 
		& -0.084 ($\pm$ 0.242)            & 0.118 ($\pm$ 0.153)          
        & 0.166 ($\pm$ 0.226)             & 0.074 ($\pm$ 0.163) 
		\\
		\midrule
		\textbf{embed. dist.} $^e$ 
		& 2.530 ($\pm$ 0.460)             & \textbf{1.499} ($\pm$ 0.195) 
        & 3.855 ($\pm$ 0.437)             & \textbf{2.891} ($\pm$ 0.351) 
		\\
		
		\bottomrule
	\end{tabular}
	\vspace{0.1cm}
	\\
	\footnotesize{
		$^a$~BLEU with smoothing \citep{lin2004orange},
		$^b$~average precision score, 
		$^c$~token ratio between generated and original content,
		$^d$~TweetEval \citep{barbieri2020tweeteval} classifications evaluated with pearson correlation coefficient (higher is better) and aggregated across subclasses,
		$^e$~pairwise embedding distance (lower is better) based on TwHIN-BERT \citep{zhang2023twhin} [CLS] token
	}
	\caption{
		Comparison of the base model - Llama-3.2-3B \citep{dubey2024llama} - (in-context prompting) and the aligned version - adapter fine-tuning \citep{yu2023low} - on our German and English Twitter politician and follower datasets for the posting and replying task evaluated on $n=100$ independent random samples for $k=10$ repetitions not seen during training. Bold marks the in-class (language, task) best values if significant by standard deviation. 
	}
	\label{tab:results:generative}
\end{table*}

%% file: figures/results.probability.test.tex
\begin{table*}[ht]
	\renewcommand{\arraystretch}{1.15}
	\centering
	\begin{tabular}{ll | ccc | r}
		\hline
		\textbf{Language} & 
        \textbf{Class} & 
        \textbf{Precision} & 
        \textbf{Recall} & 
        \textbf{F1-Score} & 
        \textbf{Support} \\
		\hline
		\multirow{3}{*}{\textbf{German}} & 
        Ignored
        & 0.741
        & 0.630
        & 0.681
        & 500
        \\
		& 
        Replied        
        & 0.678              
        & 0.780           
        & 0.726             
        & 500
        \\
        \cline{2-6}
		& 
        \textbf{Weighted Avg}   
        & \textbf{0.710}              
        & \textbf{0.705}           
        & \textbf{0.703}             
        & 1000             
        \\
		\hline
		\multirow{3}{*}{\textbf{English}} & 
        Ignored        
        & 0.973              
        & 0.972          
        & 0.972             
        & 500
        \\
        & 
        Replied        
        & 0.972              
        & 0.974           
        & 0.973             
        & 500
        \\
        \cline{2-6}
		& 
        \textbf{Weighted Avg}   
        & \textbf{0.973}              
        & \textbf{0.973}           
        & \textbf{0.972}             
        & 1000
        \\
		\hline
	\end{tabular}
	\caption{
        Replying Likelihood metrics for German and English data. TwHIN-BERT \citep{zhang2023twhin} (frozen) utilized to embed the individual samples. The train/test combinations are chosen by the iteration with the best test  F1-Score.
    }
	\label{tab:results:probability:test}
\end{table*}

%% file: sections/107--Conclusion.tex
\section{Conclusions and Future Work}
In this paper we questioned the empirical realism of generative-agent-based modeling for imitating user behavior on social networks. First, we provided a formal framework for building realistic TWONs. Second, we instantiated this framework with the purpose of mimicking user behavior based on data from $\mathbb{X}$ in English and German. Third, we benchmarked the empirical realism of agents, imitating actual users.

\subsection{Key Recommendations}
Our empirical results provide several key recommendations for conducting social simulations based on generative-agent-based modeling:

First, simulation models should be validated with respect to their empirical realism before conducting simulations. Results of a simulation should always be put into perspective to the empirical realism of all components in the simulation.

Second, simulations should be performed in the same setting in which the simulation components were fitted and validated. The stark difference between English and German performance demonstrates this. The English models performed significantly closer to their real-world counterparts and produced more stable results. Changing the setting without retraining and validation can lead to unreliable outcomes of a simulation.

Third, fine-tuning of the simulation components is required to obtain sufficient levels of empirical realism. In-context prompting of LLMs, as is done in most of the related work \cite{larooij2025large}, specifically in areas like psychology, social sciences, or media studies is often not sufficient to guarantee a simulated behavior that is close to reality.

\subsection{Future Research}
This paper is a first step towards more robust empirical research designs and protocols for studying real-world social networks by simulating users with generative-agent-based models. Although we established general formal models and initial empirical insights, further research is required. The heterogeneous nature of social networks suggests that discourse patterns can vary significantly between different communities, each with its own linguistic norms, interaction styles, and argumentation preferences \cite{gnach2017social}. This diversity presents challenges both methodologically and theoretically. Methodologically, there is uncertainty about whether LLMs are adequate to capture these variations \cite{anderson2024homogenization} or if more specialized models tailored to specific communities are needed. Theoretically, this leads to a broader inquiry about whether social media discourse's inherent heterogeneity necessitates a community-specific modeling approach rather than universal models.

%% file: sections/201--Postscript.tex
\section*{Limitations}
We have already extensively discussed the limitations of our framework and experiments throughout this paper, specifically in Sect.~\ref{subsec:ImiLimi}. What should be added is that our limited quantitative experiments do not sufficiently answer considerations regarding complex discourse quality metrics such as polarization into separate communities of users. So far, we have only simulated and analyzed one-turn communicative behavior. However, our framework also allows us to assess more sophisticated metrics that span discussion threads. 

Also, to what extent computational models capture nuanced variations in argumentation styles across different user communities would require qualitative evaluations, which we have not included in this paper. Discourse patterns can vary significantly between different communities, each with its own linguistic norms, interaction styles, and argumentation preferences. This heterogeneity presents both methodological and theoretical challenges. From a methodological perspective, we must consider whether our current modeling approaches are sufficiently sophisticated to capture these variations or if we need to develop more specialized, community-specific models. Theoretically, this opens up a broader question about the nature of social media discourse: Are social networks inherently so heterogeneous that meaningful modeling requires a community-by-community approach, rather than attempting to develop a universal model for online argumentation?

\section*{Ethical Considerations}
As is typical for most AI methods, the modeling approach presented in this paper is a dual-use technology. While social simulation - more concretely, replication of social networks and imitation of social media users - is primarily intended to be used to analyze social networks, the findings can also be used to adjust network mechanics, like ranking or filtering, which in turn influences public debate and opinion formation of users on those networks. Again, this can be used to improve debate quality and contribute to well-informed opinion formation. However, it can also be used to spread one-sided propaganda, misleading information, or manipulative advertisements.

\textbf{Privacy and Consent Considerations:} A significant ethical concern in our study involves the use of real user data from $\mathbb{X}$ to train models that simulate individual behavior patterns. Although our data set consists of publicly available posts from political figures and public replies from regular users, the individuals whose data we used did not provide explicit informed consent for their communication patterns to be replicated by generative models. This raises important questions about digital privacy rights, even when dealing with public data. The simulation of specific individuals' posting and replying behaviors creates synthetic content that mimics their communication style, potentially enabling the creation of convincing but fabricated posts that could be attributed to real people. This capability poses risks of impersonation, identity theft, misrepresentation, and potential damage to the reputation of the individuals whose behavior patterns were learned by our models.

Unfortunately, our research objectives, demonstrating "empirical realism" (aka "ecological validity") of our simulation, do not allow us to use synthetic personas rather than real user data. Creating fictional user profiles with diverse characteristics would eliminate privacy concerns, but without demonstrating the validity of the simulation. We argue that you can only claim the applicability of findings produced via social simulations to the real world once you have proven the "empirical realism" (aka "ecological validity") of your simulation. The key component for showing this is replicability on an individual level, which requires real-world data from individuals. Synthetic personas cannot be used to prove "empirical realism", but real-world data is essential to prove the empirical realism of synthetic personas. And once you prove the empirical realism of synthetic personas with real-world data, you again have a sufficiently similar replica of real-world users. Thus, you cannot have both: A valid and provable realistic social simulation without using real-world user data to model agents and measure their empirical realism.

Thus, we decided not to publish either our data set or the fitted models of individual users. We only publish aggregated statistics that do not allow one to draw conclusions on an individual user level. This hinders the replicability of results but respects individual privacy.



\newpage

\section*{Acknowledgments}
This study was conducted with a financial contribution from the EU’s Horizon Europe Framework (HORIZON-CL2-2022-DEMOCRACY-01-07) under grant agreement number 101095095.